\documentclass[traditabstract,referee]{aa}

\usepackage[colorlinks, linkcolor=black,  anchorcolor=black, citecolor=black]{hyperref}
\usepackage{hyperref}
\usepackage{subfigure}
\usepackage{graphicx}
\usepackage[varg]{txfonts}
\usepackage{longtable}
\usepackage{pdflscape}
\usepackage{hhline}
\usepackage{natbib}
\nolinenumbers
\modulolinenumbers[0]  
\renewcommand{\linenumbers}{\relax}  

\begin{document}

\title{General Methods for Evaluating Collision Probability of Different Types of Theta–phi Positioners}

\author{
Baolong Chen\inst{1}
\and Jianping Wang\inst{1}
\and Zhigang Liu\inst{1}
\and Zengxiang Zhou\inst{1}
\and Hongzhuan Hu\inst{1}
\and Feifan Zhang\inst{2}\thanks{E-mail: ffz@ahmu.edu.cn}
}

\institute{
Department of Precision Machinery and Precision Instrumentation, USTC, Hefei, People's Republic of China
\and
School of Biomedical Engineering, AHMU, Hefei, People's Republic of China
}
\date{Received date /Accepted date }

\abstract{
In many modern astronomical facilities, multi-object telescopes are crucial instruments. Most of these telescopes have thousands of robotic fiber positioners(RFPs) installed on their focal plane, sharing an overlapping workspace. Collisions between RFPs during their movement can result in some targets becoming unreachable and cause structural damage. Therefore, it is necessary to reasonably assess and evaluate the collision probability of the RFPs. In this study, we propose a mathematical models of collision probability and validate its results using Monte Carlo simulations. In addition, a new collision calculation method is proposed for faster calculation(nearly 0.15\% of original time). Simulation experiments have verified that our method can evaluate the collision probability between RFPs with both equal and unequal arm lengths.Additionally, we found that adopting a target distribution based on a Poisson distribution can reduce the collision probability by approximately 2.6\% on average.
}

\keywords{instrumentation: adaptive optics – methods: analytical – telescopes}
\titlerunning{General Methods for Calculating Collision Probability }
\authorrunning{baolongchen}
\maketitle
\section{Introduction}

Spectra of celestial bodies contain a wealth of physical information. Accurately obtaining and analyzing spectra help to revolutionize our comprehension of the formation and evolution of the universe \citep{collision_avoidance_in_next,Overview_of_the_LAMOST}. The large-scale multi-object spectroscopic facilities can significantly increase the number of simultaneously observed celestial objects \citep{an_8_mm_diameter}, thus making significant contributions to  addressing numerous pressing issues in astrophysics and cosmology \citep{collision_avoidance_in_next,Fibre_assignment_in_next}.

Fibre-fed spectrographs are being considered the most versatile spectrograph type in Multi-object facilities\citep{A_fiber_positioner_robot,collision_avoidance_in_next,an_8_mm_diameter}. A common issue for all the fiber-fed spectrographs are simultaneously positioning thousands of fiber ends with high accuracy \citep{A_fiber_positioner_robot}. To address this problem, various fiber positioning technologies have been proposed. In the Sloan Digital Sky Survey project(SDSS)\citep{Sloan_Digital_Sky_Survey_Early_Data_Release,The_Seventh_Data_Release_of_the_Sloan_Digital_Sky_Survey,The_Tenth_Data_Release_of_the_Sloan_Digital_Sky_Survey,SDSS-IV}, fibers are manually inserted into holes drilled in the focal plane plate\citep{The_2.5_m_Telescope}. Except for this method, pick-and-place devices\citep{2dF} is another method. Pick-and-place devices, used in facilities such as the Two-degree Field (2dF) facility, utilize a robot to sequentially place fibers on the focal plane. Consequently, the reconfiguration time increases linearly with the number of fibers\citep{The_Anglo_Australian_Observatory_2dF}. Thus, this method is inefficient for large telescopes with high multiplexing capabilities. 

The most efficient devices are robotic fiber positioner (RFP) arrays \citep{Comparison_of_communication}. RFP arrays can simultaneously position thousands of fibers within a very short time \citep{SDSS-V_Algorithms,Motion_Planning_for_the_Robotic}.
Currently, there are two main types of RFP arrays: the tilting spine fiber positioners and the theta-phi robotic fiber positioners. The former type is primarily employed in projects such as the Fiber Multi-Object Spectrograph (FMOS)\citep{Fibre_Multi-Object_Spectrograph} and the 4-metre Multi-Object Spectroscopic Telescope (4MOST)\citep{4MOST}. This technology eliminates the issue of fiber twisting and allows for a smaller pitch, thereby facilitating observations of closely spaced targets, approximately 1 mm apart \citep{Sphinx}. However, achieving larger tilt angles results in reduced light incidence, and the limited precision of motion necessitates multiple iterative operations to reach the desired position.

Theta-phi RFP is a kind of positioner with two degrees of freedom. RFP achieves the target position by rotating both the central and the eccentric arm via motor drives. Notably, it has been utilized by projects such as the Large Sky Area Multi-Object Fiber Spectroscopic Telescope (LAMOST)\citep{LAMOST}, the Dark Energy Spectroscopic Instrument (DESI)\citep{The_DESI_fiber_positioner_system}, the Sloan Digital Sky Survey V (SDSS-V)\citep{SDSS_V_Pioneering_Panoptic_Spectroscopy}, the Multi-Object Optical and Near-IR Spectrograph (MOONS)\citep{MOONS}, the Multiplexed Survey Telescope (MUST)\citep{Conceptual_design_of_the_optical_system}, and Subaru\citep{Prime_focus_spectrograph_(PFS)_for_the_Subaru_Telescope}. The theta-phi positioner offers advantages such as robustness, easy maintenance and fast positioning with high precision. Besides, it can realize real-time compensation for errors caused by temperature fluctuations and atmospheric refraction. Moreover, it aligns directly with celestial objects, thereby minimizing light loss \citep{Closed}. 

 Currently, RFPs are used or proposed for most focal plane systems with multiplexing ability. For the theta-phi RFP arrays, the overlapped patrol areas between adjacent RFPs ensure high coverage proportion. However, this design introduces the risk of collisions between RFPs, which can not only damage the devices but also render some critical observation targets unobservable. \citep{Galaxy_and_Mass_Assembly,galaxies_statistics_cosmology,Galaxy_kinematics_and_dynamics}. Additionally, a single RFP malfunction or disconnection can significantly affect the entire RFP array\citep{topology}.Therefore, it is necessary to analyze the probability of collisions. Analyzing collision probability can be beneficial for optimizing the design of fiber positioning systems.

In the process of RFPs reaching their targets, researchers have identified several primary causes of collisions. Due to the overlapped workspace and the physical volume of the fiber ends, collisions may occur \citep{collision_avoidance_in_next,an_8_mm_diameter}. Additionally, collisions often happen between close targets \citep{Fibre_assignment_in_next,The_Effect_of_Fiber_Collisions}. However, the collision probabilities have not been thoroughly analyzed. 

The ratio of the overlapping area to the total area can be used to approximately estimate the probability of collisions \citep{Collision_possibility_analysis_and_collision_avoidance}. However, this method does not account for the physical volume of the RFPs and the specific parameters such as safe distance in path planning. The binomial distribution has also been utilized to analyze collision probabilities, but it only considers cases with closely distributed targets\citep{an_investigation_of_collisions}. Some researchers have also analyzed the types of collisions between theta-phi positioners and calculated the probability of collisions. Nevertheless, their calculations were only for equal-arm RFPs (LAMOST scheme) and are not applicable to unequal-arm RFPs \citep{Collision_possibility_analysis_and_collision_avoidance}.

In this study, we introduce a comprehensive mathematical model for assessing the collision probabilities of theta-phi RFP arrays. This efficient, generalized approach is applicable to a broad range of existing fiber positioning systems. We validate our model using the Monte Carlo method, confirming its reliability. This work not only establishes a robust framework for evaluating collision probabilities but also provides references for the design and analysis of theta-phi RFP arrays.

This article is organized as follows: In Section 2, we introduce the fiber positioning system, providing a comprehensive overview of the existing facilities. Section 3 delves into the detailed mathematical methods used to evaluate collision probabilities. Section 4 describes the application of the Monte Carlo method in calculating the collision probabilities of RFPs. Finally, in Section 5, we compare and discuss the results obtained from the mathematical and Monte Carlo methods, providing an analysis of their respective strengths and weaknesses.

\section{Fiber Positioning System}
\label{sec:Introduction}

\begin{figure}
\centering
	\includegraphics[width=0.75\columnwidth]{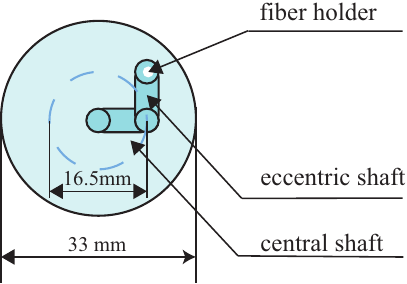}
    \caption{Schematic of equal-arm RFP. The dashed circle represents the patrol area of the RFP.}
    \label{fig:Lamost fiber positioner}
\end{figure}

RFPs are widely used in the fiber positioning system. Their structure is similar to that of SCARA robots, featuring a 2R robotic configuration. Two rotational shafts driven by two motors move the fiber end to arbitrary position within the patrol area (represented by outermost area in Fig. \ref{fig:Lamost fiber positioner}). The kinematics can be calculated based on the kinematics of a 2R robot(The calculation methods are mentioned in \citet{collision_avoidance_in_next,SDSS-V_Algorithms,Collision_possibility_analysis_and_collision_avoidance}). Thus, the patrol area is determined by the lengths of central and eccentric arms. In the focal plane, the positioners are hexagonally packed with overlapped patrol area.  

\begin{figure}
\centering
	\includegraphics[width=0.6\columnwidth]{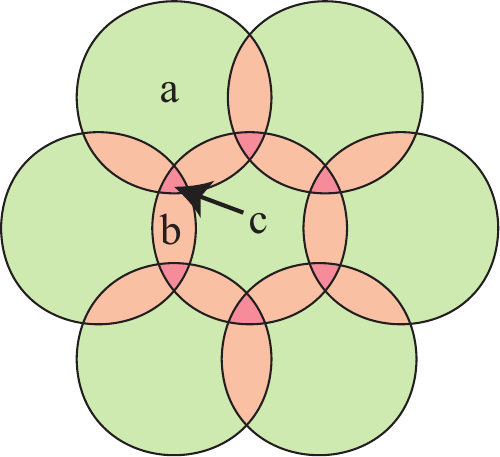}
    \caption{Accessibility of equal-arm RFPs. Region a: only one RFP can reach, region b: two RFPs can reach, region c: three RFPs can reach.}
    \label{fig:Lamost fiber positioner distribution}
\end{figure}

The differences in the extent of patrol area overlapping among various types of RFPs are primarily influenced by three parameters: pitch, central and eccentric arm length. Based on these parameters, RFP can be categorized into two main types. One type is the equal-arm RFP, i.e., the arm length of eccentric arm is equal to that of the central arm. Its patrol area is shown in Fig.\ref{fig:Lamost fiber positioner distribution}. In the overlapped areas represented by b and c, the eccentric arm of neighbouring RFPs may collide with each other, damaging themselves. This design is currently adopted by projects such as LAMOST, DESI, and PFS\citep{Instrumentation_at_the_Subaru_Telescope,DESI,Prime_focus_spectrograph_(PFS)_for_the_Subaru_Telescope}.

\begin{figure}
\centering
	\includegraphics[width=0.7\columnwidth]{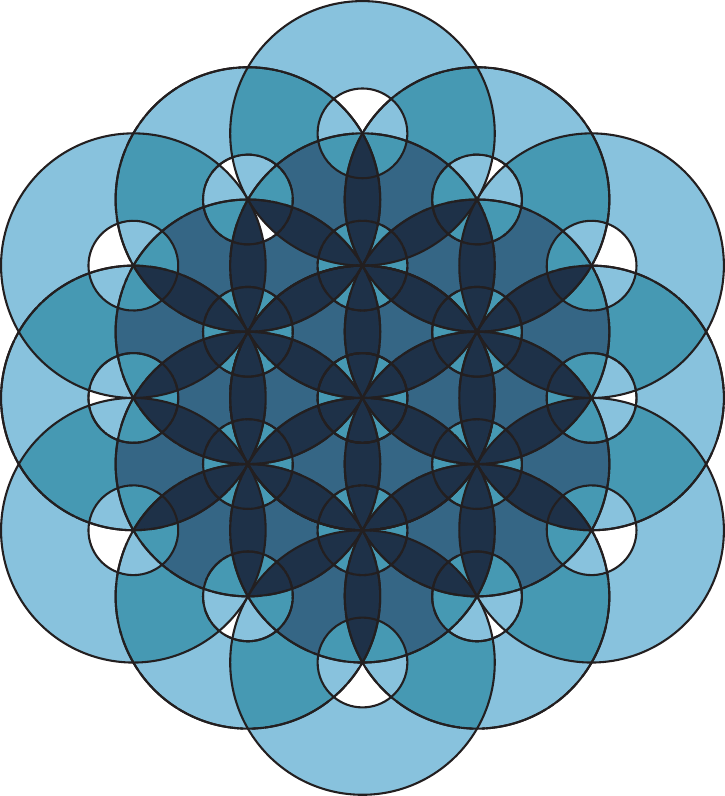}
    \caption{Accessibility of unequal-arm RFP in different regions. The shading in the figure indicates the coverage of each region by the RFPs. The depth of the color indicates the coverage of a single area by the RFP: the darker the color, the more RFPs can reach that area.}
    \label{fig:SDSS fiber positioner alloca}
\end{figure}

Another type is positioners with unequal arms. For unequal-arm positioners, the length of eccentric arm is usually longer than that of the central arm. Thus, an unequal-arm positioner cannot reach its own center, forming a ring-shaped area. For this design, certain regions can be accessed by multiple positioners, thereby enhancing the overall coverage of the focal plane by the RFP arrays. This improvement directly contributes to a higher coverage proportion, a key parameter for evaluating the coverage of the focal plane by RFP arrays. However, this design also increases collision probability. The schematic of overlapped patrol areas of densely packed unequal-arm positioners is shown in Fig. \ref{fig:SDSS fiber positioner alloca}.  Currently, this design is used or prepared to be used in projects such as SDSS-V and MOONS. 

\section{Mathematical Models of Collision Probability}
\label{sec:Mathematical}
The collisions of the RFPs can be categorized into dynamic and static collisions. The dynamic collisions occur during the movement process. The static collisions occur when reaching the target. These two collision typies will be analyzed and discussed separately.
\subsection{dynamic collisions}
\label{sec:dynamic collisions}

The probability of dynamic collisions depends on the path planning algorithm employed and the collision caused by actual hardware failures. Various methods have been developed to diminish dynamic collision probability.

Researchers have proposed various path planning algorithms to effectively resolve dynamic collisions. Specifically, researchers in SDSS-V project propose two schemes to solve the collisions between unequal arm positioners: the greedy heuristic and the Markov chain variant of the greedy heuristic, achieving mean efficiency of approximately 99.2\% and >99.9\%, respectively \citep{SDSS-V_Algorithms}. Additionally, solutions based on the RRT algorithm for equal-arm RFPs have achieved a success rate of 99.9\% \citep{Motion_Planning_for_the_Robotic}. Furthermore, various path planning schemes based on a novel decentralized navigation function \citep{collision_avoidance_in_next} and solutions combining artificial potential field with control layers \citep{Priority_coordination_of_fiber} have been proposed to address collision issues.

Although simulation results of path planning algorithms suggest that the collision problem is nearly solved, real path planning remains complex. For instance, in the path planning method proposed for MOONS, the RFPs may exhibit creep behavior during movements.\citep{moons_path}. Additionally, during operation, collisions still occur within the fiber positioning systems due to factors such as mechanical failure. 

According to the data from the LAMOST closed-loop operation test, even with the assistance of software, 2.5\% of the RFPs still fail to reach their intended targets. Additionally, some of them may experience collisions due to various factors. Thus, hardware-based obstacle avoidance strategies are also adopted to protect RFPs. From a hardware perspective, measures such as installing sensors in the RFPs are employed to prevent dynamic collisions. These sensors are triggered when RFPs get close to each other, causing them to stop or retreat, thereby preventing a collision and protect RFPs from damage \citep{Preliminary_design}. This indicates a potential dynamic collision.

Each fiber positioning system has its own characteristics. Additional data is needed to accurately assess the dynamic collision probabilities. In this analysis, we will assume a dynamic collision probability of zero, with the option to incorporate LAMOST data into results if deemed necessary.

These abovementioned solutions primarily address dynamic collisions that occur during the movement of RFPs to their targets. However, they do not consider collisions when reaching targets, referred to as static collisions.

\subsection{Static collisions}
Static collisions cannot be fully resolved through tiling and target allocation because these stages do not fully account for the constraints of RFPs. Moreover, ignoring static collisions results in incomplete observation, causing some important targets to be abandoned during the path planning process. It is crucial to develop a method for calculating the probability of static collisions. Here, we propose a general method to evaluate the collision probability of different types of RFPs, considering both target distribution and the constraints of RFPs.

Since the equal-arm RFP can be regarded as a special case of the unequal-arm RFP (with an arm length ratio of 1), our analysis will primarily focus on collisions between unequal-arm RFPs. In this analysis, we first analyze the different types of static collisions. Subsequently, we propose a general method to calculate the probabilities for static collisions.

To accurately model the RFP array in mathematical models, it is essential to clarify the meanings of several variables. Specifically, the variable \( \text{pitch} \) refers to the spacing between adjacent RFPs, and\( l_1 \) and \( l_2 \) represent the central and eccentric arm lengths, respectively. Furthermore, the adjacent RFPs are denoted as \( \text{fiber}_n \). Additionally, \( N \) denotes the number of positioners that may potentially collide, while the arm length ratio of positioners is represented by \( \text{ratio}_i \).

When calculating the distance between positioners, the collision threshold(or safe distance), becomes critical for collision detection and path planning. This threshold is referenced from the SDSS project. The minimum safe distance \( D_{ij} \) between RFPs is then calculated as:
\begin{equation}
    \min {D_{ij}} > {\rm{MD}} + 2d
    \label{eq:min_distance}
\end{equation}
where \( i \) and \( j \) denote the indices of adjacent RFPs. A collision is considered to occur if the calculated minimum distance between RFPs is less than the specified safe distance (for example, ${D_{ij}}=5.078 mm$ for SDSS-V). The maximum displacement limit of the eccentric arm during a single-step operation (represented by \( MD \)) is determined by:
\begin{equation}
\centering
    {\rm{MD}} = (l_1 + l_2) \sin (2\Delta \theta)
    \label{eq:md_calculation}
\end{equation}
where \( l_1 \) and \( l_2 \) represent the lengths of the central and the eccentric arm, respectively. The angle \( \Delta \theta \) is controlled by the program. During a single-step operation, MDs are managed accordingly. 

In this context, it is crucial to note that the safe distance \( d \) is a key parameter. This parameter sets the relative size of the beta arm, a radial distance from the beta arm line segment. Proper setting of this parameter is vital for the path smooth operation and reliability of the fiber positioners. Considering only static collisions,  \( \Delta \theta \) is set to zero for the analysis.

Based on the pitch between RFPs, relationship between RFPs can be preliminarily divided into three types:
\begin{figure*}
\centering
\begin{tabular}{@{}c@{}}
\includegraphics[height=3cm]{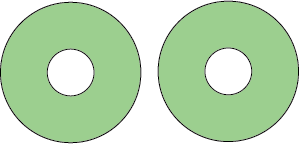} \
\textbf{(a)}
\end{tabular}

\vspace{0.02\textwidth}

\begin{tabular}{@{}c@{}}
\includegraphics[height=3cm]{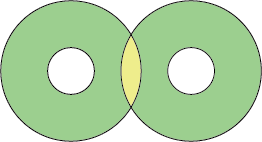} \
\textbf{(b)}
\end{tabular}

\vspace{0.02\textwidth}

\begin{tabular}{@{}c@{}}
\includegraphics[height=3cm]{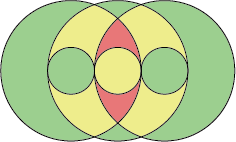} \
\textbf{(c)}
\end{tabular}

\caption{Comparison of three different collision cases.
(a) Type 1: Two RFP do not collide with each other.
(b) Type 2: The RFP experiences collisions limited to its
immediate neighboring ring of positioners.
(c) Type 3: This type of collision involves the RFP's collision
with RFP more than its neighbor positioners.
In the images, the green region represents the area covered by a single RFP, the yellow region is covered by two RFPs, and the red region is covered by three RFPs.}
\label{fig:comparison}
\end{figure*}

\textbf{Type 1:} The positioner does not collide with its neighboring RFPs, \( N = 0 \). In this case, the condition \( \text{pitch} > 2 (l_1 + l_2 + d) \) needs to be satisfied, as shown in the Fig. \ref{fig:comparison}a. 

\textbf{Type 2:} The positioner collides with the neighboring ring of RFP, \( N \leq 6 \). In this case, the condition \( 2 \text{pitch} > 2  (l_1 + l_2 + d) > \text{pitch} \) needs to be satisfied, as shown in the Fig. \ref{fig:comparison}b. For equal arm RFPs, this can be simplified to the distribution map of LAMOST and the calculation method mentioned in Feifan Zhang's paper\citep{Collision_possibility_analysis_and_collision_avoidance} can be used.

\textbf{Type 3:} This type of collision involves the RFP's collision with RFP more than its neighbor RFPs, \( N > 6 \). In this case, the condition \( 2  (l_1 + l_2 + d) > 2  \text{pitch} \) needs to be satisfied, as shown in the Fig. \ref{fig:comparison}c.

\subsection{Collision Calculation}
We propose a new method for simultaneously calculating the collision probabilties of Type 2 and Type 3. Unlike the previous method that only calculated the overlapping area, the proposed approach divides the collision probability into two parts. They are the ratio of the conflict area (i.e., overlapped area) to the total coverage area and the second the ratio of the collision area of the eccentric arm of RFPs to the coverage area (i.e., patrol area). Thus, the collision probability is approximated by the following equation: 

\begin{figure}
\centering
	\includegraphics[width=\columnwidth]{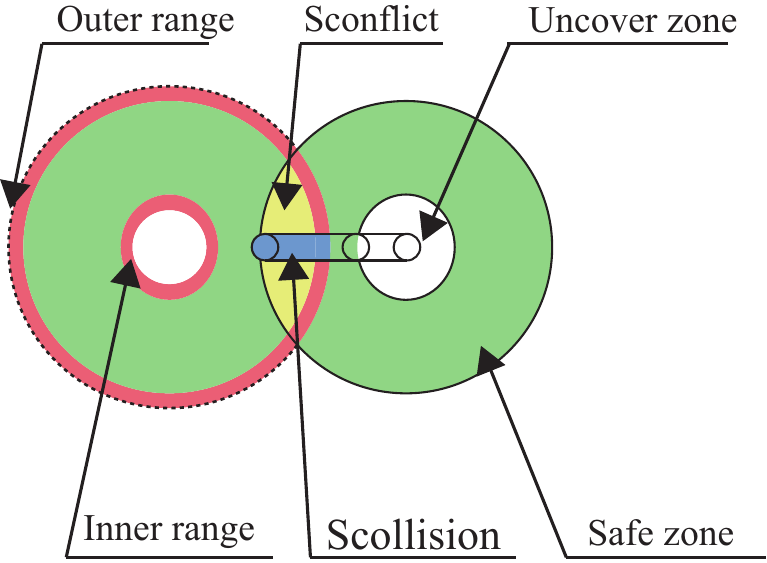}
    \caption{The diagram illustrating the calculation of collision probability. The green, blue and yellow areas represent the patrol area of one positioner, the potential collision zones of the eccentric arm, and the conflict area, respectively. The red areas represent the actual inner and outer diameters considering obstacle avoidance.}
    \label{fig:ppc}
\end{figure}

\begin{equation}
\centering
\label{equ:collision p}
P = \sum_{i=0}^{N} \left( \frac{S_{\text{conflict},i}}{S_{\text{cover}}} \times \frac{S_{\text{collision},i}}{S_{\text{cover}}} \right)
\end{equation}

 where \(S_{\text{conflict}}\) is the conflict area between positioners (yellow  part in Fig. \ref{fig:ppc}) .\(S_{\text{cover}}\) is the coverage area of positioners, and \(S_{\text{collision}}\) represents the overlapping area of the maximum motion ranges of the eccentric arm and another fiber positioner's motion range, which is denoted as blue in Fig. \ref{fig:ppc}. To account for the impact of the Collision Threshold \( D_{ij} \), the calculation range of the RFP needs to be slightly expanded. \(S_{\text{collision}}\) and \(S_{\text{cover}}\) can be calculated by Eqs. \ref{equ:collision s} and \ref{equ:cover s}.
 
\begin{equation}
\centering
\label{equ:collision s}
S_{\text{collision}} = d \times( l_2+d) \times \max\left( 0, \frac{l_1 + l_2 - pitch/2 }{l_2/2} \right)
\end{equation}

\begin{equation}
\centering
\label{equ:cover s}
S_{\text{cover}} = \pi \left( R_{\text{outer}}^2 -  R_{\text{inner}}^2 \right)
\end{equation}
where \(R_{\text{outer}}\) and \(R_{\text{inner}}\) denotes the maximum and minimum radius of the motion circle (Fig. \ref{fig:ppc}) calculated by Eqs. \ref{equ:inner} and \ref{equ:outer}.

\begin{equation}
\centering
\label{equ:inner}
R_{\text{inner}} = l_2 - l_1 - d
\end{equation}

\begin{equation}
\centering
\label{equ:outer}
R_{\text{outer}} = l_2 - l_1 + d
\end{equation}

\section{Collision Probability Calculation Using Monte Carlo Method}
\label{sec:Fiber} 
\subsection{Collision detection of fiber positioner}
\label{sec:Collision calculation for fiber positioners}
\begin{figure}[h]
\centering
	\includegraphics[width=\columnwidth]{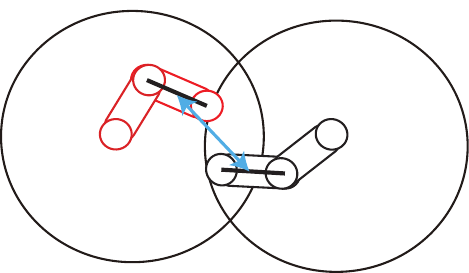}
    \caption{Distance computation between RFPs. Black thick line represents the simplified segment of the RFP. Blue arrowed line segment represents the distance between the RFP.}
    \label{fig:distance between fibers}
\end{figure}

Collisions between RFPs primarily arise due to collisions between the arms. However, since the central and the eccentric arm are not in the same plane, we only need to consider collisions between eccentric arms. Excluding the zero-position design of baffles found in some RFPs and the curvature of the eccentric arm\citep{Test_results_of_the_SDSS-V,LAMOST_Fiber_Positioning_Unit_Detection_Based_on_Deep_Learning}, the eccentric arm can be simplified to a straight line segment, as shown in Fig.\ref{fig:distance between fibers}. We can determine whether a collision occurs by calculating the shortest distance between the eccentric arms of adjacent RFPs. If the calculated distance is less than the collision threshold, a collision happened\citep{SDSS-V_Algorithms}. Thus, the distance calculation is crucial for collision detection of thousands of RFPs. Here, we propose a new method for collision detection. We calculate the distance between eccentric arms by discretizing the lines into pairs of points. The calculation accuracy depends on the number of discrete segments. Here, we divide the lines into 64 segments. In order to improve the computation efficiency,  these distances are computed by CUDA framework based on GPU parallel computing.

Comparison with the distances calculated by Python standard package Scipy indicates that the proposed method satisfies the accuracy requirements for path planning. Compared to currently used method of LAMOST (i.e., detecting collisions by calulating sweeping areas of positioners)\citep{Collision_free_motion_planning}, the proposed method can significantly improve the computation speed. Utilizing CUDA parallel computation with 4090D GPU, the computation time of collision detection for 4,000 RFPs is reduced from approximately 300 seconds on the current system to nearly 0.45 seconds. This substantial reduction in computation time greatly accelerates the iteration speed of optimization algorithms. Also, this helps to improve the reconfiguration efficiency of close-loop controlled fiber positioning systems by increasing the path planning efficiency.

\subsection{Collision probability calculation}
\subsubsection{Tiling and target allocation}

The calculation of collision probability must depend on the existing observation methodologies such as target allocation algorithms because the collision probabilities correlates with target distribution and allocation. Here, we briefly introduce the current tiling strategy and target allocation algorithms.

Before each observation, schemes including the pointing of telescope and the star catalogs must be determined. For LAMOST, researchers have developed a Sky Strategy System (SSS) to generate observation plates with the goal to fully utilize LAMOST's multiplexing capabilities. For most sky areas, SSS adopts the maximum density algorithm, density gradient algorithm, and mean shift algorithm to select targets for each observation. Each target in the input star catalog is assigned a priority based on the scientific needs. Each spectrograph reserves 5 flux standard stars and 20 sky fibers for data processing\citep{A_network_flow_algorithm_to_position_tiles_for_LAMOST,The_first_data_release_(DR1)_of_the_LAMOST_regular_survey}.

For SDSS project, a network flow algorithm is used for Tile. The placement of observation fields adopts an iterative optimization strategy. It starts with uniformly distributed observation fields. Then, their positions are adjusted to minimize the target allocation cost. After that, target will be reallocated and observation fields will be updated. This process will be iteratively executed until minimum cost is obtained. Finally, a binary search is used to determine the minimum number of observation fields to meet SDSS requirements\citep{An_Efficient_Algorithm_Positioning_Tiles, An_Efficient_Targeting_Strategy}.

After the allocation of plates or observation fields, targets are allocated to positioners. For each observation of LAMOST, SSS assigns targets from the input star catalog to positioners according to  their priority.  If multiple targets with identical priority are within one positioner's patrol area, the one nearest to positioner is selected so as to minimize collision risks with adjacent fibers. If a target is accessible by multiple positioners, it will be allocated to the RFP with the lowest current target count. A total of 20 RFPs without target allocated will be selected as the sky fibers \citep{Observation_planning_of_LAMOST_fiber_positioning_subsystem_and_its_simulation_study}. In cases where fewer than 20 sky fibers are allocated to a spectrograph, fibers with the lowest-priority targets are reassigned as sky fibers until the required number of sky fibers are achieved. Note that the number of sky fibers varies with surveys. For the BOSS observations, some fibers need to be reserved for sky observation \citep{The_Eighteenth_Data_Release}. Based on the design and scientific requirements of SDSS-V RFPs, researchers propose an algorithm to allocate the telescope's RFPs to various types of targets, considering factors such as priority, exposure time, target characteristics, and dynamic targets \citep{The_SDSS-IV_MaNGA_Sample,Heterogeneous_Target_Assignment_to_Robotic_Fiber_Positioner_Systems}.

\begin{figure*}
  \centering
  \includegraphics[width=\textwidth]{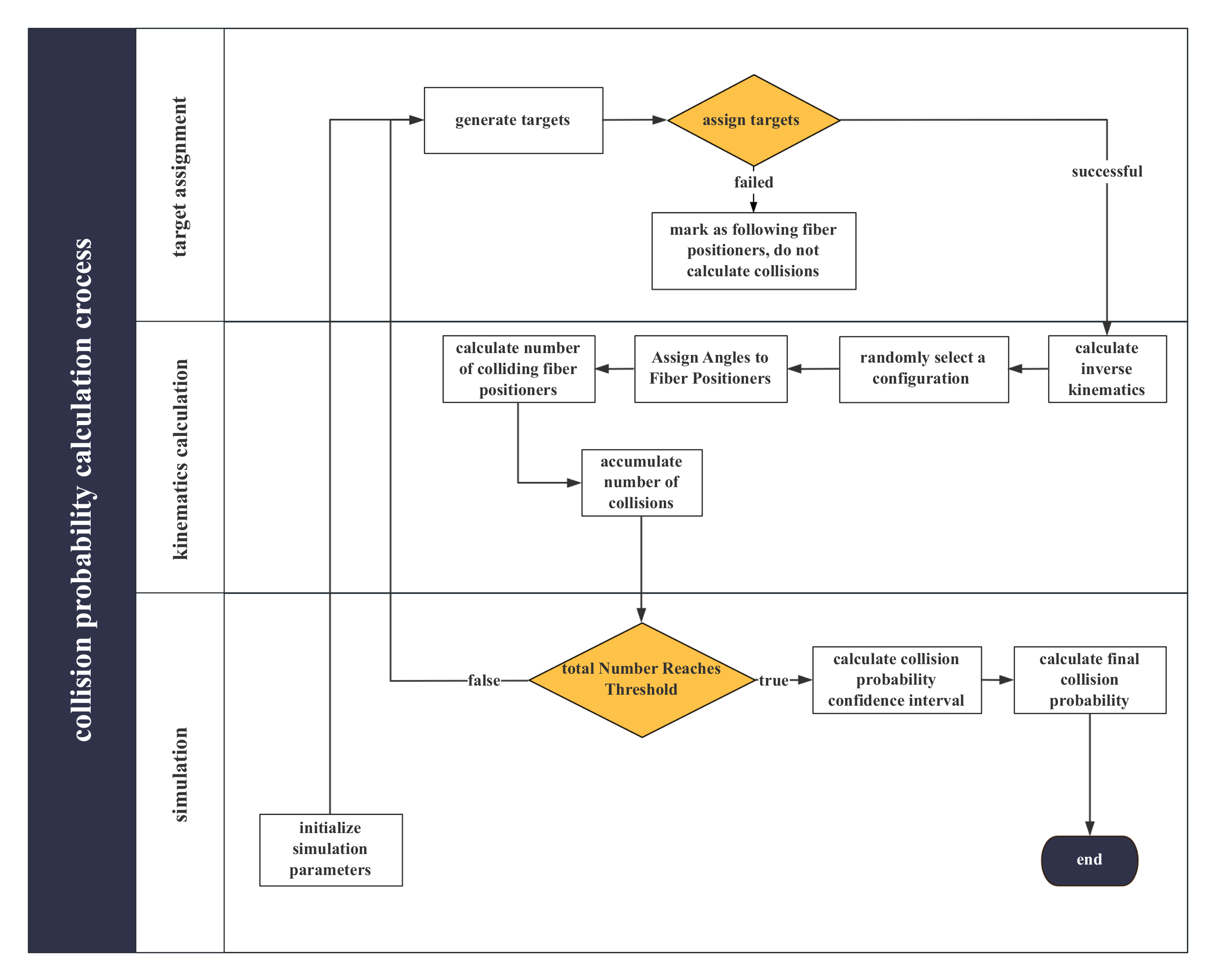}
  \caption[Collision Prob. Calc. Flow]{Collision probability calculation flowchart.}
  \label{fig:flowchart}
\end{figure*}

\subsubsection{Monte Carlo simulation process of fiber positioners}
Compared to previous methods, the Monte Carlo approach requires extensive simulations. To facilitate these simulations while ensuring accuracy, some modifications are proposed.

When validating the mean shift algorithm, the NSNO 2.0 star catalog is adopted,  whereas SDSS validated its algorithm based on Cole's work \citep{Simulating_the_formation_of_cosmic_structure}. However, since this study focuses on simulating the distribution of targets on the focal plane for multiple observations of the same tiling, we adopt simulated catalogs:  Poisson distributed targets employed by Feifan Zhang and uniformly distributed targets from LAMOST surveys \citep{Fiber_assignment_for_multi-object_fiber-fed_spectrographs,Fiber_assignment_in_wide-field_multiobject}. We neglect the requirements for sky background subtraction,bright stars, and the priorities of the targets, as these factors vary across different projects and observations. These needs can be met through additional settings and adjustments in large-scale optimizations. Furthermore, we do not exclude targets that are too close to each other because removing some targets results in the decreasing of simulated collision possibility, deviating the real one  \citep{A_convergent_mean_shift_algorithm_to_select_targets_for_LAMOST,Target_allocation_yields_for_massively_multiplexed_spectroscopic_surveys,Collision_possibility_analysis_and_collision_avoidance}.

\begin{figure}
\centering
	\includegraphics[width=\columnwidth]{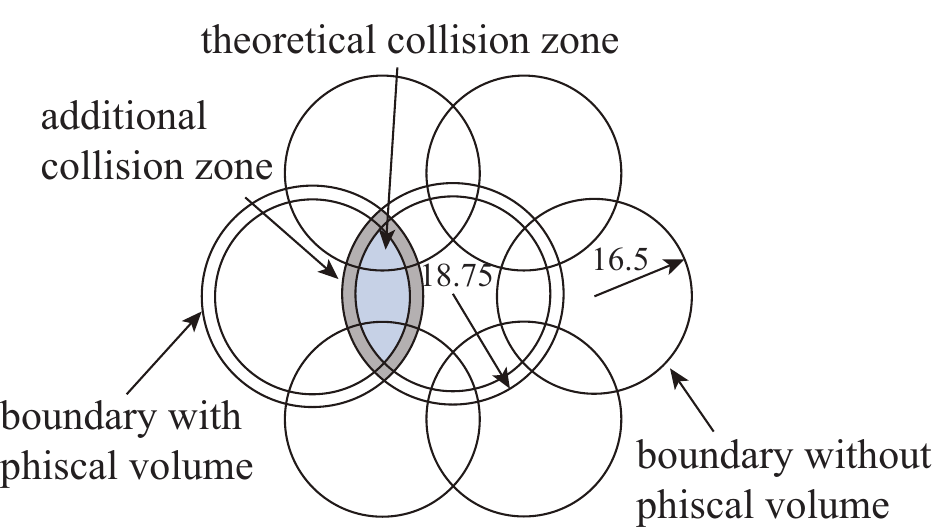}
    \caption{This distribution map shows seven RFP. The larger circles represent the actual movement areas, while the smaller circles denote the theoretical movement areas of the RFP. The light blue area indicates the theoretical conflict zone, which excludes the physical volume of the fiber heads. The gray area accounts for the additional physical volume needed to avoid collisions of the fiber positioner's eccentric shaft. The RFP spacing is 25.6 mm, leaving little remaining safety area.}
    \label{fig:lamost collision region}
\end{figure}

In this study, we aim at calculating the collision probability of all positioners operating at full capacity. To achieve this goal, it is necessary to consider that different spectroscopic surveys have various target-to-fiber ratios. For instance, the ratios for LAMOST, DESI, and SDSS are 4, 5, and 1, respectively. In this scenario, 20,000 randomly generated targets are distributed within an area slightly larger than the focal plane. New targets will be generated for each simulation.

Then targets will be allocated to positioners with fewer current targets and closer positioners. Thus nearly every robot will be assigned more than one target. This target allocation strategy is simple. It avoids the influence of complex allocation strategies on collisions and simplifies the simulations.In addition, subsequent optimization of the allocation strategy can be adopted to reduce collision probability.

In scenarios where RFP have not been assigned a target, for equal-arm positioners, one approach, as adopted by LAMOST, is to keep them in their original positions. However, collision may occur. Calculations indicate that the actual safe zone is quite limited, as shown in Fig.\ref{fig:lamost collision region}. For unequal-arm positioners, there are no fixed safe positions or safe angles. Thus, these positioners should be designated as following positioners. They will act as auxiliaries in path planning and collision avoidance, yielding space for other positioners. Thus, their collisions are ignored and motion planning algorithm will be responsible for their collision avoidance.

Based on the reasons mentioned in section\ref{sec:dynamic collisions} , we will therefore disregard the dynamic collision probability, which has a minor impact in practice. Thus, we will treat the collision probability as equivalent to the static collision probability. We do not consider the target loss arising from unsolvable path planning problems and hardware failures(impact is minimal). The overall calculation flowchart is shown in Fig.\ref{fig:flowchart}

\subsubsection{Number of iterations of Monte Carlo Simulations}
The precision of Monte Carlo simulations arises with the number of iterations. However, increasing the number of iterations leads to a higher computational cost. Thus, a proper number of iterations is needed. There are generally two methods to determine the number of iterations required for Monte Carlo simulations. One is based on the convergence criterion and the other relies on confidence intervals and confidence levels. As the collision probability follows a skewed distribution\citep{Topology_Recoverability_Prediction_for_Ad-Hoc_Robot_Networks}, the first method of calculating collision probability based on the convergence criterion is insufficient. To address this challenge, we employ the Wilson confidence interval method, renowned for its robustness and reliability when dealing with skewed distributions. The Wilson method elegantly computes the upper and lower bounds of the confidence interval using equations \ref{eq:wilson confidence1} and \ref{eq:wilson confidence2}, effectively providing a more reliable estimate of the collision probability. 

\begin{equation}
\centering
U = \frac{1}{1 + \frac{z_{\alpha / 2}^2}{n}} \left( \hat{p} + \frac{z_{\alpha / 2}^2}{2n} + z_{\alpha / 2} \sqrt{\frac{\hat{p}(1 - \hat{p})}{n} + \frac{z_{\alpha / 2}^2}{4n^2}} \right)
\label{eq:wilson confidence1}
\end{equation}

\begin{equation}
\centering
L = \frac{1}{1 + \frac{z_{\alpha / 2}^2}{n}} \left( \hat{p} + \frac{z_{\alpha / 2}^2}{2n} - z_{\alpha / 2} \sqrt{\frac{\hat{p}(1 - \hat{p})}{n} + \frac{z_{\alpha / 2}^2}{4n^2}} \right)
\label{eq:wilson confidence2}
\end{equation}

Here, \(\hat{p}\) represents the proportion of collisions, \(n\) denotes the number of Monte Carlo simulations, and \(z\) is the z-statistic corresponding to a given confidence level. The confidence level is set to be 0.95. The mean of the upper and lower bounds is taken as the collision probability. Simulations demonstrate that the method based on confidence intervals and levels is truly more stable than the convergence criterion. And the proper number of iterations is 6000.

\section{Methods Validation of Mathematical Model}
\label{sec:Methods}
\subsection{Validation in Small-scale Positioners}

To validate the mathematical model, we take 19 RFPs as an example. The parameters for the Monte Carlo method are presented in Table \ref{tab:Parameters and their values for the fiber positioner simulation}. In this analysis, we vary two variables: pitch and arm length ratio. This analysis focuses on a preliminary examination of how collision probability changes with the parameters of the RFP. This serves as a reference for subsequent designs.

Based on the RFPs currently adopted by LAMOST, the length of the eccentric arm can be easily adjusted by exchanging it, thereby altering the arm length ratio. By referring to the design schemes of other RFPs, the arm length ratio range is designed to be between 1 and 3. Currently, the pitch between RFPs is 25.6 mm.In the simulation, this variable is designed to range between 25.6 mm and 35 mm.

\begin{table}
\centering
\caption{Parameters and their values for the fiber positioner simulation}
\label{tab:Parameters and their values for the fiber positioner simulation}
\begin{tabular}{lc}
\hline
Parameter & Value \\
\hline
Number of fiber positioners & 19 \\
Monte Carlo iterations & 6000 \\
Base arm length & 8.25 mm \\
arm ratio range & 1--3 \\
Pitch & 25.6--35 mm \\
Target distribution range & circle with a 153.6 mm diameter \\
Safe distances & 4.5mm \\
\hline
\end{tabular}
\end{table}

\begin{figure*}
  \centering
  \begin{tabular}{@{}ccc@{}}
    \includegraphics[width=0.45\textwidth]{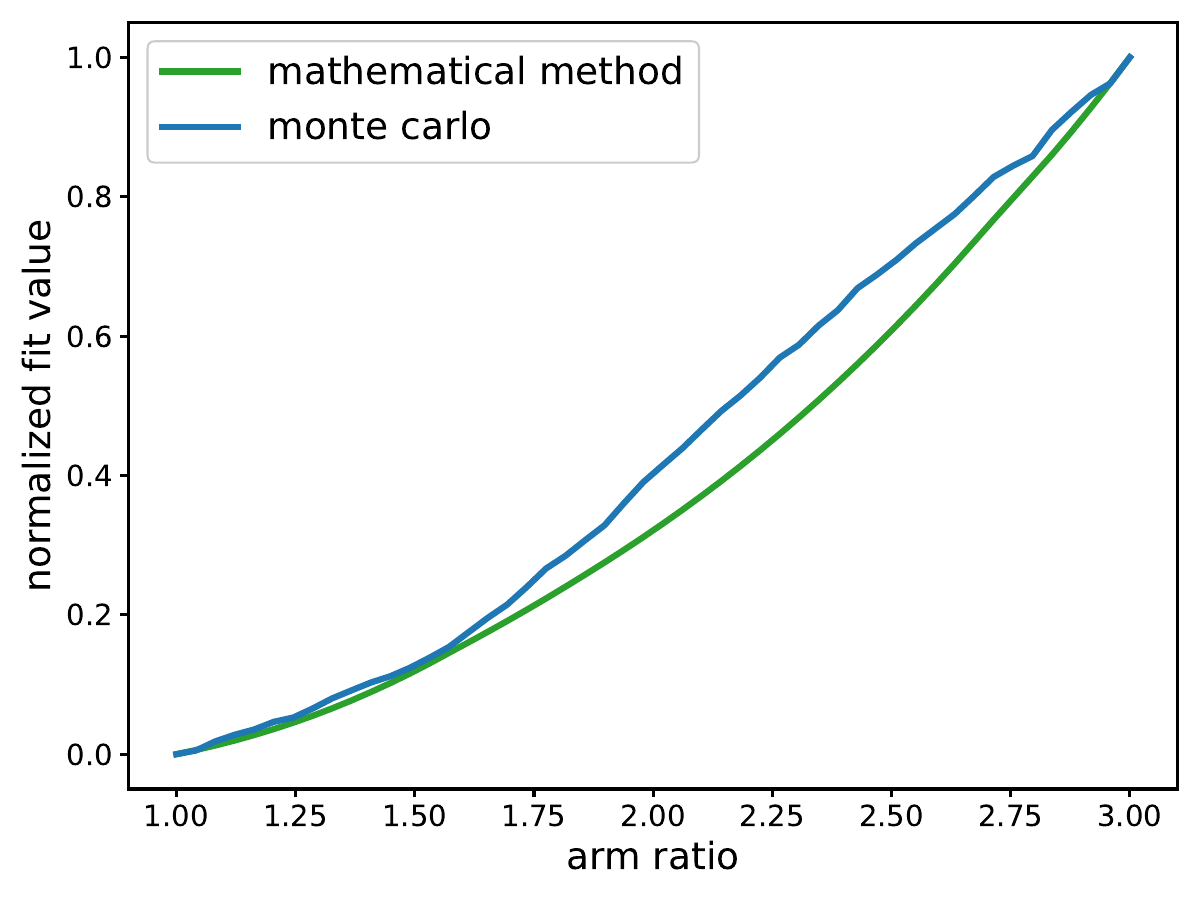} &
    \includegraphics[width=0.45\textwidth]{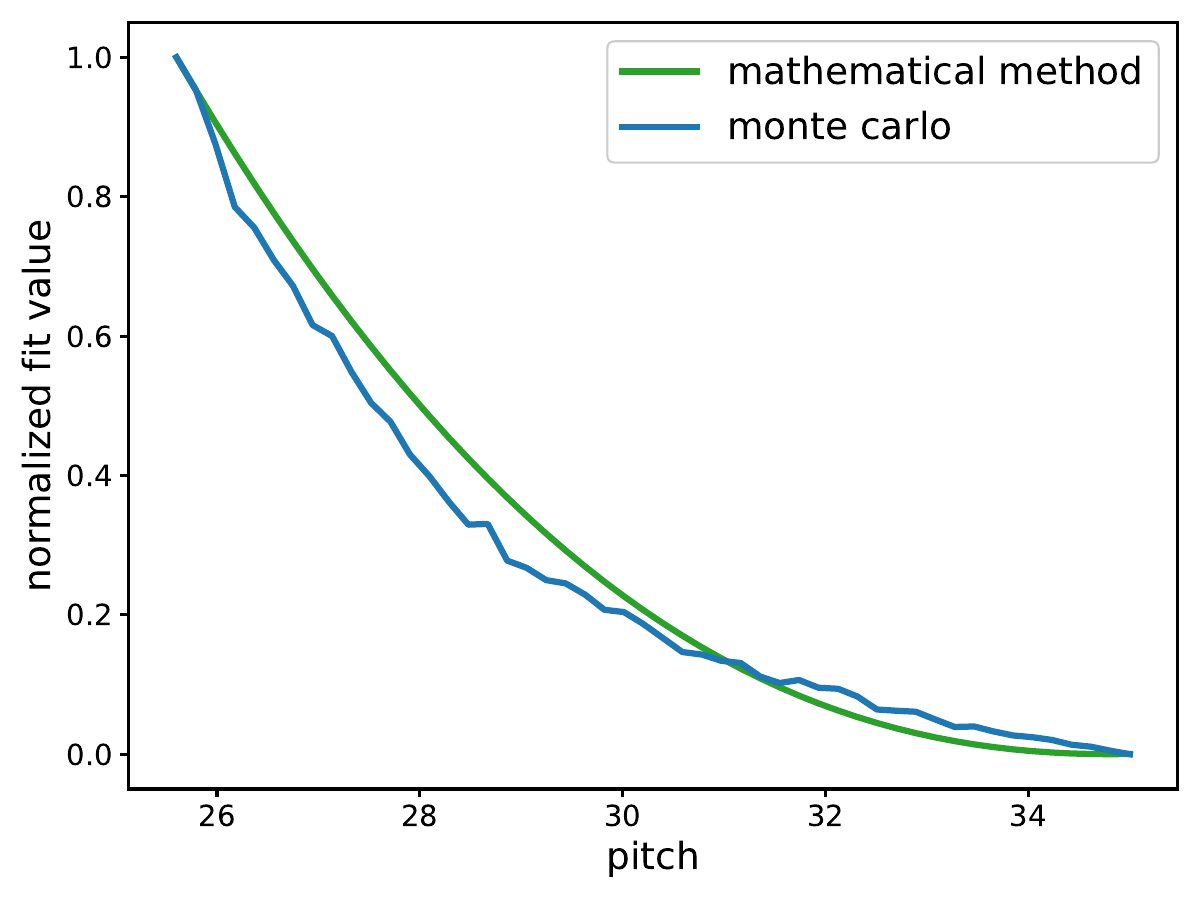} \\
    \textbf{(a)} & \textbf{(b)} 
  \end{tabular}
\caption{(a) The relationship between the arm ratio and collision probability
(b) The relationship between the pitch and collision probability. In each plot, the green line represents the results from the proposed calculation method, and the blue line represents the results from the Monte Carlo simulations.}
  \label{fig:comparison1}
\end{figure*}

\begin{figure*}
\centering
\begin{tabular}{@{}c@{}}
\includegraphics[width=0.48\textwidth]{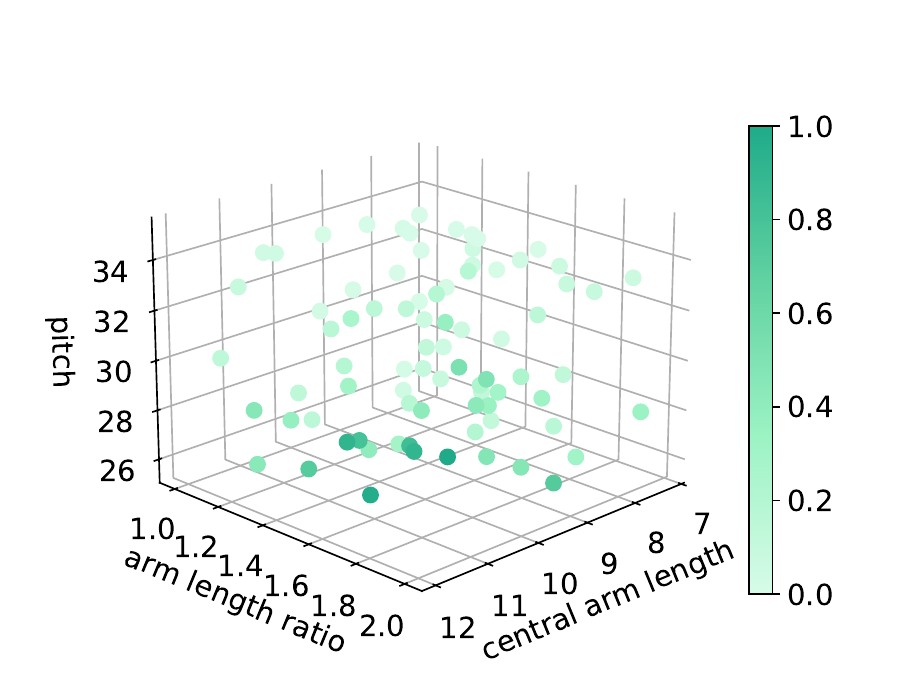} \
\textbf{(a)} \
\includegraphics[width=0.48\textwidth]{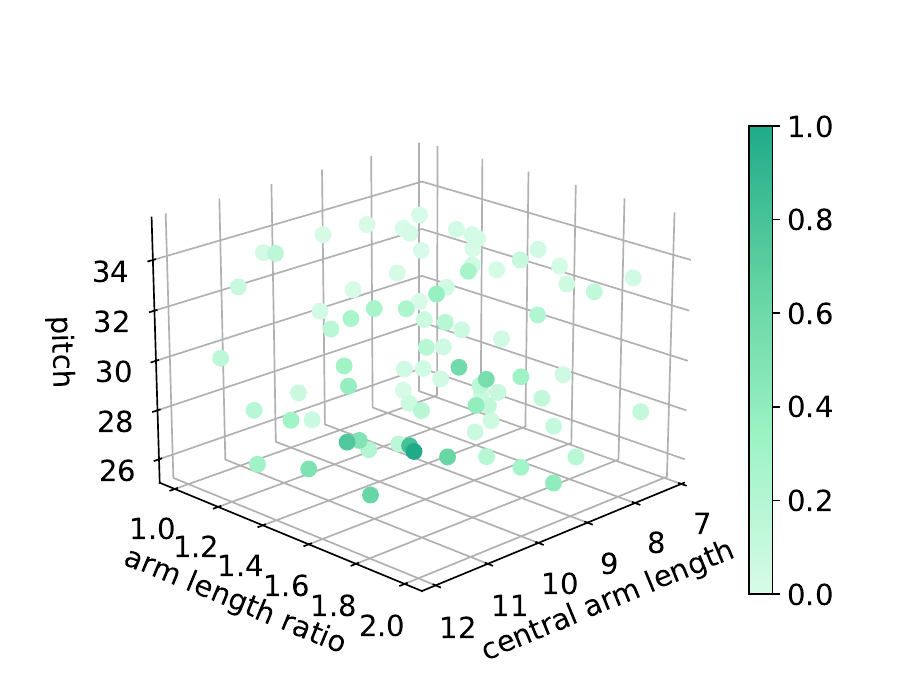} \
\textbf{(b)}
\end{tabular}
\caption{Results of different methods. (a) The Monte Carlo method, (b) The Mathematical calculation method. In each plot, the color depth represents the normalized collision probability, with darker colors indicating higher collision probabilities.}
\label{fig:comparison2}
\end{figure*}

\begin{figure*}
\centering
\begin{tabular}{@{}c@{}}
\includegraphics[width=0.48\textwidth]{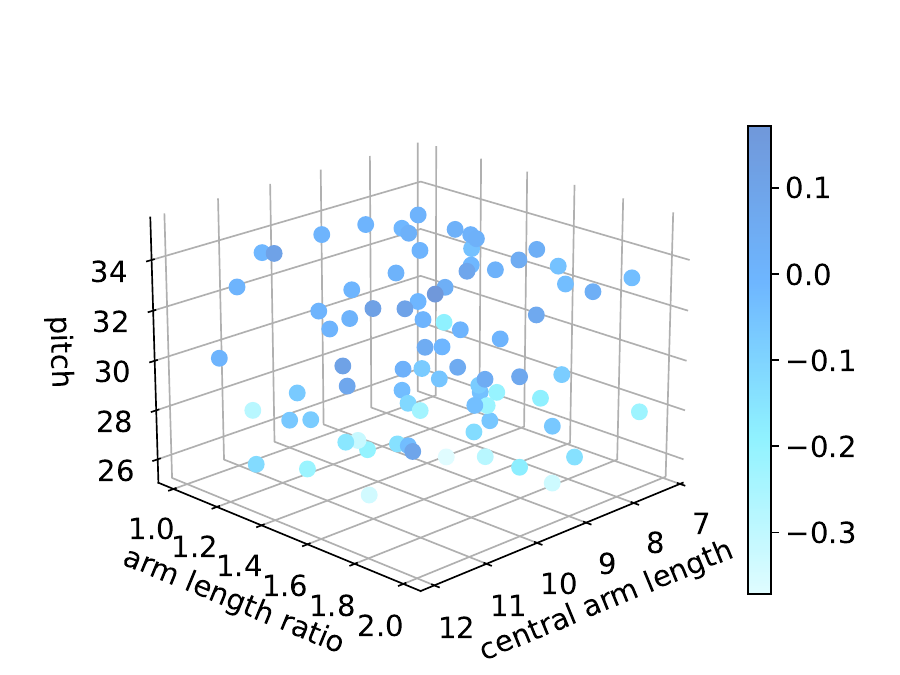} \
\textbf{(a)} \
\includegraphics[width=0.48\textwidth]{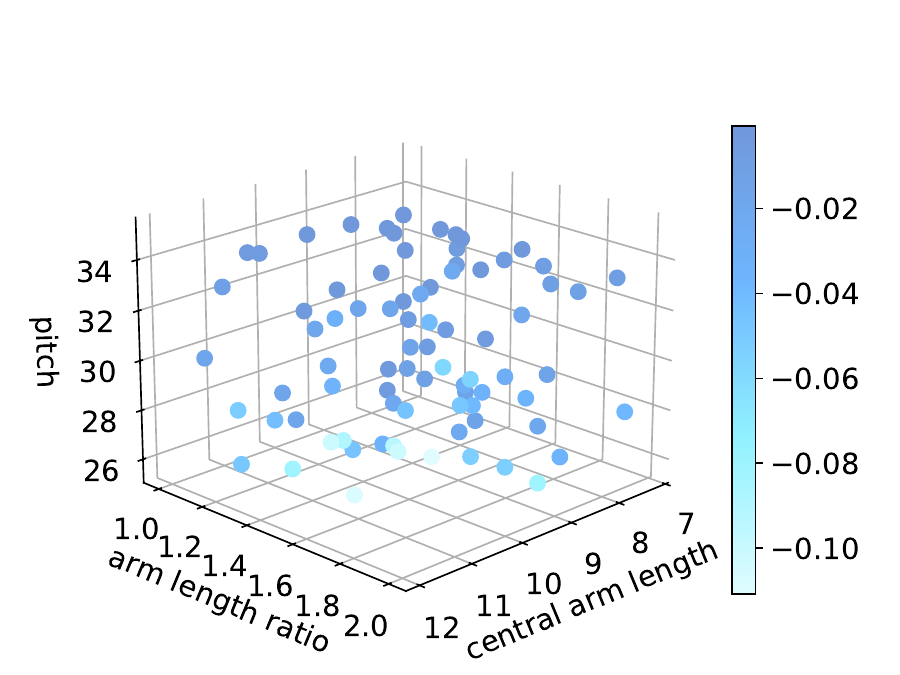} \
\textbf{(b)}
\end{tabular}
\caption{Comparison of results obtained from different methods. (a) Residuals between the Monte Carlo method  and the mathematical method(results are normalized)
(b) Residuals between the Monte Carlo method using uniform distribution and the Monte Carlo method using Poisson distribution(original data). In each plot, the color depth represents the difference between results.}
\label{fig:comparison3}
\end{figure*}

We compared two methods: the proposed mathematical method and Monte Carlo method. The verification results are shown in Fig. \ref{fig:comparison1}. To enable comparison across different scales, the results of two methods were normalized. These results indicate that the two methods yield comparable outcomes. The proposed mathematical method can effectively evaluate collision probability, with only minor variations arising from Monte Carlo simulation fluctuations. Additionally, slight discrepancies in results are attributable to the use of substitutions, and estimates in the calculations. Nevertheless, these differences are minimal and do not significantly impact the overall evaluation effect, making the method sufficiently robust for evaluation purposes.

Moreover, it is evident that the collision probability increases rapidly with the increase in arm ratio, while it decreases rapidly with the increase in pitch. This is significant for design purposes, as it can help determine how to compromise and adjust multiple parameters.

Although a correlation has been observed, it is difficult to reach a generalizable conclusion due to the small scale of the positioner arrays. Therefore, larger-scale simulation experiments are needed to further verify this relationship. 

\subsection{Validation in Large-scale Positioners}

To obtain more general results, we increased the number of RFPs and allowed the vary lengths of both arms, as well as the pitch, to vary within certain ranges. Some parameters have been appropriately adjusted to accommodate large-scale RFP arrays. We use parameters in Table\ref{tab:parameter_settings}.

\begin{table}
    \centering
    \caption{Parameter Settings for Large-scale Positioners}
    \begin{tabular}{ll}
        \hline
        Parameter & Value \\
        \hline
        Monte Carlo iternations & 6000 \\
        Collision threshold (mm) & 4.5 \\
        Number of positioner Holes & 469 \\
        Optimization Variables & Arm Length Ratio, Pitch \\
        Pitch Range (mm) & 24.6-35 \\
        Arm length ratio & 1-2\\
        Arm Length Range & 7.25-14.5 \\
        Target distribution range & 200\\
        Target num & 20000\\
        \hline
    \end{tabular}
    \label{tab:parameter_settings}
\end{table}

In this study, we randomly distributiond eighty data sets within the variable range and applied two computational methods: the proposed Mathematical Model and the Monte Carlo Method. For ease of comparison, we normalize the results. The results are shown in the Fig.\ref{fig:comparison2}. Furthermore, to investigate the impact of the distribution of targets on the focal plane, we utilized two different distributions: Poisson distribution and uniform distribution. This aims to assess how the distribution form of targets affects efficiency.

After performing normalization on the data, we observe that their overall distribution shapes are similar in Fig.\ref{fig:comparison2}. Nevertheless, differences in details persist, particularly when the central arm length and arm ratio are significantly large. 

We use the Monte Carlo method as a benchmark to better observe these differences. The results obtained using the proposed method are then subtracted from the benchmark for comparison. The mean difference between the results of the two methods is -0.050, with a variance of 0.116. Fig.\ref{fig:comparison3} suggests that the mathematical calculation method is highly correlated with the Monte Carlo method in evaluating collision probabilities. However, a significant discrepancy between these methods arises when the arm length ratio is large and the pitch is increased. This discrepancy may arise due to the limited range and specific target distributions utilized in the Monte Carlo method. As the pitch increases, the patrol area of some RFPs may extend beyond the scatter range, leading to the distortion in collision probability calculations. This phenomenon also occurs when calculating collision probability using Monte Carlo method in small scales.

 Additionally, for a clearer comparison between different distributions, we directly subtract the collision probabilities of the Poisson distribution and the uniform distribution, yielding the results shown in the Fig.\ref{fig:comparison3}. We note that in Fig.\ref{fig:comparison3}, the collision probability of the Poisson distribution is consistently lower than that of the uniform distribution. The average collision probability for a uniform distribution is 2.6\% higher compared to a Poisson distribution. Moreover, as the pitch and arm length ratio increases, the difference between the two gradually becomes larger. This provides a new approach to reducing collision probability in the fiber positioning system. When tiling, a certain distribution of targets can be selected to reduce collision probability.

To facilitate future applications, we fit data and establish the relationship between three independent variables: arm length (\(x\)), arm length ratio (\(y\)), and pitch (\(z\)), with the dependent variable being the collision rate (\(f\)).

The functional form of the model is:

\begin{equation}
\centering
\label{eq:fit}
f(x, y, z) = a + bx + cy + dz + ex^2 + fy^2 + gz^2 + hxy + ixz + jyz
\end{equation}

\begin{displaymath}
f(x, y, z) = a + bx + cy + dz + ex^2 + fy^2 + gz^2 + hxy + ixz + jyz
\end{displaymath}

where \(a, b, c, d, e, f, g, h, i, j\) are the equation parameters to be estimated through data fitting.

Using a Polynomial Regression model with Regularization, the parameters were estimated, as presented in Table\ref{tab:model_parameters}. The model demonstrates a strong goodness of fit, achieving a high R-squared value of 0.934 on the test set. This robust model lays a solid foundation for subsequent RFP design and collision probability evaluations.

\begin{table}
\centering
\caption{Estimated model parameters .}
\label{tab:model_parameters}
\begin{tabular}{cr}
\hline
Parameter & Estimated Value($10^{-3}$) \\
\hline
$a$ & $18.6635$ \\
$b$ & $4.61227$ \\
$c$ & $15.9225$ \\
$d$ & $-15.0398$ \\
$e$ & $1.07078$ \\
$f$ & $2.67496$ \\
$g$ & $2.87948$ \\
$h$ & $2.23372$ \\
$i$ & $-3.19570$ \\
$j$ & $-8.26024$ \\
\hline
\end{tabular}
\end{table}

\section{Discussions}

Collisions between RFPs are inherently complex. While numerous algorithms offer path planning solutions, these solutions heavily depend on hardware reliability and communication systems. For example, the decentralized navigation function has been proposed for DESI project to solve collisions. However, the planned path must be discretized due to the limitation of communication systems. After discretization, collision avoidance is not guarateed. For large-scale arrays, a high collision probability necessitates sophisticated path planning algorithms and system reliability. 

Accurate assessment of collision probabilities is essential for evaluating the effectiveness of fiber positioning systems in telescopes. By integrating collision analysis during the design phase, it is possible to select configurations that minimize collision probability. This proactive approach not only enhances the system's efficiency but also alleviates the operational burden on both the software and hardware components.

This article introduces a novel method for evaluating the collision probability of theta-phi RFPs, which is applicable to all current designs. Although this method exhibits a slightly larger error compared to the approach proposed by Feifan Zhang, it is versatile, accommodating both positioners with equal arm lengths and those with unequal arm lengths. 

The proposed mathematical calculation method offers several distinct advantages over the Monte Carlo method. Most notably, it provides a significantly faster calculation speed. In addition, the results are not influenced by either the target distribution range or the distribution itself, which enhances the robustness and reliability of the method. 

However, the Monte Carlo-based approach discussed in this paper remains a valuable method for calculating collision probabilities. This method allows for the calculation of various types of target distributions and different kinds of RFPs. Additionally, it accounts for the influence of tile and target allocation on collision probabilities, thereby providing a more comprehensive analysis.

Furthermore, collision detection methods in the Monte Carlo-based method have significantly reduced the computational time. Now, analyzing the collisions of 4000 RFPs takes just 0.45 seconds. This is pivotal for real-time path planning and replanning.

Our analysis reveals that employing the Poisson distribution results in a lower estimated collision probability (average 2.6\% Fig.\ref{fig:comparison3}). This finding suggests a novel approach to reducing collision probability by assigning targets with varying distribution forms to the fiber positioning system. For instance, when tiling, selecting a specific distribution of targets could effectively reduce the probability of collisions.

The landscape of survey-based astronomy is progressing rapidly, largely driven by the continuous development of highly automated telescope and instrument systems. Although previous studies have studied various aspects of collision calculation, the collision probability associated with different types of RFP arrays has not been fully explored. Furthermore,  collision between RFPs represent just one facet of survey control. Optimizing target allocation, survey scheduling can also enhance the overall efficiency of astronomical surveys.

\section{Summary}
\label{summary}

Spectroscopic surveys play a critical role in addressing numerous pressing questions in astronomy and astrophysics. Most large-scale multi-object surveys employ theta-pi positioners, aiming to simultaneously place the fiber ends precisely at their target locations. To avoid blind spots, the patrol areas of these positioners overlap, potentially leading to collisions. We propose a general model to calculate the probability of these collisions. Monte Carlo simulations have been used to verify our results. The analysis revealed a mean difference of -0.05 between the two methods, with an associated variance of 0.116.

Furthermore, we examined two target distributions and analyzed their respective collision probabilities. Our analysis indicates that by accounting for the distribution forms of targets, we can reduce collision probabilities by an average of 2.6\%.  In future studies, we will explore how these collision probability evaluation methods can be employed to optimize the parameter design of RFPs.

\begin{acknowledgements}
Funding for this research was provided by Cui Xiangqun’s Academician Studio. Guoshoujing Telescope [the Large Sky Area Multi-Object Fiber Spectroscopic Telescope (LAMOST)] is a National Major Scientific Project built by the Chinese Academy of Sciences. Funding for the project was provided by the National Development and Reform Commission. LAMOST is operated andmanaged by the National Astronomical Observatories, Chinese
Academy of Sciences.
This work was supported by the the Engineering Science Experimental Teaching Center, University of Science and Technology of China was acknowledged to provideinstruments and rooms.And we thank the USTC Supercomputing Center for computing resource.

\end{acknowledgements}


\bibliography{reference}
\bibliographystyle{aa}

\appendix

\end{document}